\documentclass[runningheads]{llncs}
\usepackage{graphicx}

\usepackage{tikz}
\usepackage{comment}
\usepackage{amsmath,amssymb} 
\usepackage{color}

\usepackage{lipsum}
\newcommand\blfootnote[1]{%
  \begingroup
  \renewcommand\thefootnote{}\footnote{#1}%
  \addtocounter{footnote}{-1}%
  \endgroup
}

\begin{document}
\pagestyle{headings}
\mainmatter

\track{LVIS}
\title{A Good Box is not a Guarantee of a Good Mask} 

\titlerunning{A Good Box is not a Guarantee of a Good Mask}
%
\author{Jingru Tan\inst{1*} \and
Gang Zhang\inst{2*} \and
Hanming Deng\inst{3*},
\\
Changbao Wang\inst{3},  Lewei Lu\inst{3}, Quanquan Li\inst{3}, Jifeng Dai\inst{3}}
\authorrunning{J. Tan et al.}
%
\institute{Tongji University \and
Tsinghua University \and
SenseTime Research
}
\maketitle
\blfootnote{*\text{Equal Contribution}}
\begin{abstract}
This article introduces the solutions of the team \textit{lvisTraveler} for LVIS Challenge 2020. In this work, two characteristics of LVIS dataset are mainly considered: the long-tailed distribution and high quality instance segmentation mask. We adopt a two-stage training pipeline. In the first stage, we incorporate EQL and self-training to learn generalized representation. In the second stage, we utilize Balanced GroupSoftmax to promote the classifier, and propose a novel proposal assignment strategy and a new balanced mask loss for mask head to get more precise mask predictions. Finally, we achieve 41.5 and 41.2 AP on LVIS v1.0 val and test-dev splits respectively, outperforming the baseline based on \text{X101-FPN-MaskRCNN} by a large margin.

\end{abstract}



 
 \section{Introduction}
 
 LVIS is a new dataset for large vocabulary instance segmentation. Firstly, given modern object detectors perform poorly in few samples regime, it provides new research opportunities for long-tailed object detection. Secondly, unlike COCO dataset \cite{lin2014microsoft}, it provides over 2 million high quality mask annotations, making it possible to train and evaluate against high quality ground truth.
 
 Our solutions focus on those two aspects: (1) handling the extremely inter-class imbalance caused by long-tail distribution, (2) predicting higher quality instance mask. Overall, we adopt a two-stage training strategy consisting of the representation learning stage and the fine-tuning stage. At the representation learning stage, we use some techniques like EQL \cite{tan2020equalization}, repeat factor re-sampling \cite{gupta2019lvis}, data augmentation, self-training to learn generalized representation. At the fine-tuning stage, we first freeze the backbone, and follow the balanced group softmax to balance the classifier for solving the inter-class imbalance problem. We also put more emphasis on the mask head at this stage. We found that a well-aligned bounding box does not guarantee a precise mask. For example, instances of some categories usually have large bounding boxes but with thin masks. i.e, the area ratio of the mask and bounding box is small. However, given a proposal, the traditional strategy is to extract features at a specific feature map according to scale of the bounding box, as a consequence, the required detailed information for predicting thin mask may be discarded at the coarse feature map if a proposal has large bounding box. To alleviate this problem, we assign the mask proposals considering both scale of the bounding box and area ratio of the mask and bounding box. Another issue caused by the extremely small area ratio is the imbalance problem of foreground and background pixels when training the mask head. So we propose a novel balanced mask loss, which combines dice loss \cite{milletari2016v} with weighted binary cross-entropy loss. Specifically, the new mask loss will dynamically adjust the weight for foreground pixels according to the area ratio.
 
 \section{Our Approach}
 
 \subsection{Representation Learning Stage}
 
 \textbf{EQL.} We apply Equalization Loss \cite{tan2020equalization} to alleviate the suppression to rare and common categories.
 
 \noindent \textbf{RFS.} Repeat Factor Sampling \cite{gupta2019lvis} is adopted.
 
 \noindent \textbf{Data Augmentation.} Mosaic \cite{bochkovskiy2020yolov4}, rotate, scale jitter is used unless otherwise stated.
 
 \noindent \textbf{Self-training.} 
 We do inference on LVIS v1.0 training data, and collect pseudo labels of bounding boxes that do not have overlap with ground truth. We consider those pseudo labels as missing annotations (caused by sparse annotation of a federated dataset). Then we ignore proposals if those proposals have a large IOU overlap with these pseudo labels. We also do inference on Open Image data, and use the pseudo labels to jointly train with standard LVIS v1.0 training data. We only sub-sample 10k images from those pseudo labels each training epoch and use a loss weight $\lambda$ to control its effect.
 
 \subsection{Fine-tuning Stage}
 
 \noindent \textbf{Classifier Balance.} After the representation learning stage, we freeze the backbone, neck, and RPN. Balanced GroupSoftmax \cite{li2020overcoming} is used for balancing classifier.
 
 \noindent \textbf{Mask Proposal Assignment.}  Unlike COCO, which includes 80 well-defined categories, the LVIS dataset has 1203 categories found by data-driven object discovery, and instances of some categories may have irregular shapes. As a result, some new challenges arise. We find instances of some categories have large scale bounding boxes but with thin masks, in other words, the area ratio of mask and bounding box is small. But in the proposal assignment stage, we usually assign proposals to specific feature map (e.g. P2, P3, P4, P5 when using FPN \cite{lin2017feature}) to extract features according to the scale of the bounding box. As a result, some proposals with large bounding boxes and thin masks will be assigned to the coarse feature map in which the required detailed information needed for predicting thin masks may be discarded. To alleviate this problem, we propose a new proposal assignment strategy for mask proposals, which considers both scale of the bounding box and area ratio of the mask and bounding box. Specifically, we assign proposals according to Eq.\ref{Assign}, where $S_{bbox}$ and $S_{mask}$ represent area of the bounding box and mask respectively, and 3 is the level index of the feature map with the coarsest resolution.
 \begin{equation}
    \label{Assign}
     assign\_level = min\lbrace\lfloor\frac{S_{bbox}}{56^2}\rfloor, \lfloor\frac{S_{mask}}{0.25 \cdot S_{bbox}}\rfloor, 3\rbrace
 \end{equation}
 
 \noindent \textbf{Balanced Mask Loss.}
 As mentioned above, instances of some categories have large bounding boxes but with thin masks, which also results in the imbalance between the foreground and background pixels during training. So we propose a new balanced mask loss as Eq.\ref{MaskLoss} to handle this problem. It consists of dice loss \cite{milletari2016v} and weighted binary cross-entropy loss.

\begin{equation}
    \label{MaskLoss}
    \mathcal{L}_{Mask}\left(p_{m}, y_{m}\right)=\mathcal{L}_{\text {Dice}}\left(p_{m}, y_{m}\right) + \mathcal\lambda \mathcal{L}_{WBCE}\left(p_{m}, y_{m}\right),
\end{equation}
where $p_m \in R^{H\times W}$ denotes the predicted mask for a particular category, $y_m \in R^{H\times W}$ denotes the corresponding mask ground truth, H and W are height and width of the predicted mask map respectively. $\lambda $ is a hyper-parameter to adjust the weight of weighted binary cross-entropy loss. We set $\lambda$ as 1 in all experiments.

\noindent Dice loss is given as follows.

\begin{equation}
    \mathcal{L}_{\text {Dice }}\left(p_{m}, y_{m}\right)=1-\frac{2 \sum_{i}^{H \times W} p_{m}^{i} y_{m}^{i}+\epsilon}{\sum_{i}^{H \times W}\left(p_{m}^{i}\right)^{2}+\sum_{i}^{H \times W}\left(y_{m}^{i}\right)^{2}+\epsilon},
\end{equation}
where $i$ denotes the $i$-th pixel and $\epsilon $ is a smooth term to avoid zero division. We set $\epsilon$ as 1 in all experiments. By the way, using the dice loss as mask supervision alone is worse than the standard binary cross-entropy loss.

\noindent Weighted binary cross-entropy loss is given as follows.
\begin{equation}
    \mathcal{L}_{WBCE }\left(p_{m}, y_{m}\right)=\sum_{i}^{H \times W} w_{m}^{i} [y_{m}^{i} logp_{m}^{i} + (1-y_{m}^{i})log(1-p_{m}^{i})]
\end{equation}

\noindent Weight for each pixel is given as follows.
 \begin{equation}
    \label{Ratio}
    w_{m}^{i} = \left\{
             \begin{array}{lr}
             1  , & y_{m}^{i}=0 \\
            \text{max}\lbrace 1, 0.5 \cdot \frac{S_{bbox}}{S_{mask}} \rbrace, & y_{m}^{i}=1 
             \end{array}
        \right.
 \end{equation}
 As Eq.\ref{Ratio} shows, when pixel $i$ is a foreground pixel, we adjust its weight according to area ratio 
 of the mask and bounding box of the proposal which the pixel $i$ belongs to.
 
  \noindent \textbf{Boundary Supervision.} In addition, we also add intermediate boundary supervision to improve mask localization accuracy following \cite{chengwhl20}.
 
   \noindent \textbf{More Computation on Head.} We also add three more convolutions for the mask head, and use the deformable RoI pooling to extract features for proposals in the second stage.
 
 \section{Experiments}
 
 \subsection{Dataset}
 
 \noindent \textbf{LVIS.} We perform experiments on LVIS v1.0 dataset \cite{gupta2019lvis}, which contains 1203 categories. It consists of 100k training images and 19.8k validation images. Note the LVIS is the \textbf{only source of training data with annotations.} We also reported our results on 19.8k \texttt{test-dev} set.
 
 \noindent \textbf{Open Image} We only use images without annotations of Open Images \cite{Kuznetsova_2020} to generate pseudo labels.
 
 \subsection{Implementation Details}

We re-implement the Mask-RCNN \cite{He_2017_ICCV} and HTC \cite{chen2019hybrid} following the origin paper. All the hyper-parameters are kept unchanged except we set weight decay to 0.00005 instead of 0.0001 for large models. The Learning rate is set to 0.02, batch size is 16 (one image per GPU). For HTC  model, we do \textbf{NOT} include the semantic segmentation branch because coco stuff annotation is not permitted. We train model with small backbone, e.g. ResNet-50 \cite{he2016deep} with 24 epoch, with learning rate divided by 10 at the 16th and 22th epoch. For large model, we train with 15 epoch, with learning rate divided by 10 at the 11th and 14th epoch. All models are initialized with ImageNet pre-trained model. 
 
 \subsection{Ablation Studies}
 
 We choose R50-FPN-MaskRCNN \cite{He_2017_ICCV} as our baseline model and mask head is class-specific, scale-jitter is adopted. Some useful enhancement techniques are shown in Table \ref{tab:representation_1}. With those methods, we improve the AP from 19.2 to 33.2. Based on this strong baseline, we then apply other methods, results are shown in Table \ref{tab:representation_2}.

 \begin{table*}
   \centering
   \setlength\tabcolsep{5pt}
   \begin{tabular}{c c c c c c | c c c c c}
      DA & EQL & RFS & HTC & S101\textsuperscript{\textdagger} & Tricks & AP & AP\textsubscript{\textit{r}} & AP\textsubscript{\textit{c}} & AP\textsubscript{\textit{f}} & AP\textsubscript{\textit{bbox}} \\
      \hline
       & & & & & & 19.2 & 1.1 & 17.0 & 29.5 & 19.9 \\
      \checkmark & & & & & & 20.3 & 1.9 & 18.8 & 29.9 & 21.1 \\
      \checkmark & \checkmark & & & & &  22.4 & 5.2 & 22.6 & 29.6 & 23.4 \\
      \checkmark & \checkmark & \checkmark &  & & & 26.2 & 17.1 & 26.2 & 30.2 & 27.0 \\
      \checkmark & \checkmark & \checkmark & \checkmark & &  & 28.8 & 19.2 & 29.1 & 32.8 & 31.3 \\
      \checkmark & \checkmark & \checkmark & \checkmark & \checkmark &  & 32.0 & 20.6 & 32.8 & 36.3 & 35.0 \\
       \checkmark & \checkmark & \checkmark & \checkmark & \checkmark & \checkmark & 33.2 & 23.7 & 33.7 & 36.8 & 36.1 \\
   \end{tabular}
      \caption{Ablation studies on LVIS v1.0 \texttt{val} set. Models are either Mask-RCNN or HTC w/o semantic branch. DA: Mosaic, rotate; EQL: Equalization Loss, RFS: Repeat Factor Sampling; HTC: Hybrid Task Cascade; S101: ResNeSt101 \cite{zhang2020resnest}. Tricks: set weight decay as 5e-5, make sampling probability in mosaic align with RFS, and make rotated bounding box align with rotated mask.
      {\textdagger}: we also add 400-1400 multi-scale training and DCN \cite{dai2017deformable} when using ResNeSt101.
      }
   \label{tab:representation_1}
\end{table*}

The results of fine-tuning stage are present at Table \ref{tab:representation_2}. First, fine-tune with Balanced GroupSoftmax improves the AP from 33.2 to 34.7, with a 3.0 gap between bounding boxes and masks. With our proposed high quality mask method, we shrink gap to 2.3 and further improve the AP to 36.1.

 \begin{table*}
   \centering
   \setlength\tabcolsep{5pt}
   \begin{tabular}{c c c c c c | c c c c c | c}
      ST & MS & S200\textsuperscript{\textdagger} & S269 & GS & HM & AP & AP\textsubscript{\textit{r}} & AP\textsubscript{\textit{c}} & AP\textsubscript{\textit{f}} & AP\textsubscript{\textit{bbox}} & $\Delta_{bbox-mask}$ \\
      \hline
       & & & & & & 33.2 & 23.7 & 33.7 & 36.8 & 36.1 & 2.9\\
      \checkmark & & & & & & 33.7 & 25.0 & 34.3 & 36.9 & 36.7 & 3.0\\
       & \checkmark & & & & & 33.9 & 23.3 & 34.3 & 38.0 & 36.2 & 2.3 \\
      \checkmark & \checkmark & \checkmark &  & & & 36.0 & 26.0 & 36.4 & 40.0 & 38.6 & 2.6\\
      \checkmark & \checkmark & \checkmark & \checkmark & &  & 36.5 & 24.8 & 37.1 & 40.8 & 39.2 & 2.7 \\
      \hline
       &  &  &  & \checkmark&  & 34.7 & 26.6 & 34.8 & 38.1 & 37.7 & 3.0  \\
      &  &  &  & \checkmark & \checkmark & 36.1 & 28.8 & 35.8 & 39.8  & 38.4 & \textbf{2.3} \\
      \hline
      \checkmark & \checkmark & \checkmark & \checkmark & \checkmark & \checkmark & 38.8 & 28.5 & 39.5 & 42.7  & 41.1 & 2.3\\
   \end{tabular}
      \caption{Ablation studies on LVIS v1.0 \texttt{val} set. Here baseline model is the best model in Table \ref{tab:representation_1}. ST: self-training with Open Image data; MS: Mask Scoring \cite{huang2019mask}; S200: ResNeSt200; {\textdagger}: use pseudo label of LVIS as ignore ground truth and add instaboost data augmentation. S269: ResNeSt269. GS: finetune with Balanced GroupSoftmax; HM: proposals assignment strategy  + balanced mask loss + boundary supervision.
      }
   \label{tab:representation_2}
\end{table*}

We apply multi-scale testing as Test Time Augmentation. We make several modifications on standard multi-scale testing. (1) We limit the valid bounding boxes range for each resolution. i.e. we only accept small bounding boxes on high-resolution images or large bounding boxes on small resolution images. (2) We slightly increase the score of rare categories when merge detected boxes from multiple scales. (3) We use standard NMS with a threshold 0.7 followed by Soft NMS \cite{liu20201st}.  (4)
We extract mask predictions from different resolution images according to scale of the bounding box and area ratio of the mask and bounding box. With these changes, we achieve the \textbf{single model result of 41.5 AP} on LVIS v1.0 \texttt{val} set.

\section{Final Results}

We submit our results on \texttt{test-dev} to the LVIS v1.0 evaluation server. The results are shown at Table \ref{tab:test_dev}.

 \begin{table*}
   \centering
   \setlength\tabcolsep{5pt}
   \begin{tabular}{l c | c c c c }
       method & eval.set & AP & AP\textsubscript{\textit{r}} & AP\textsubscript{\textit{c}} & AP\textsubscript{\textit{f}} \\
      \hline
       baseline by host & \texttt{val} & 27.26 & 19.47 & 26.13 & 31.95 \\
       ours & \texttt{val} & \textbf{41.5} & \textbf{30.0} & \textbf{41.9} & \textbf{46.0}  \\
       \hline
       baseline by host & \texttt{test-dev} & 26.86 & 20.41 & 24.9 & 31.97 \\
       ours & \texttt{test-dev} & \textbf{41.23} & \textbf{31.93} & \textbf{40.40} & \textbf{46.35} \\
   \end{tabular}
      \caption{Comparison of baselines provided by host with our method.
      }
   \label{tab:test_dev}
\end{table*}

\bibliographystyle{splncs04}
\bibliography{egbib}

\begin{thebibliography}{10}
\providecommand{\url}[1]{\texttt{#1}}
\providecommand{\urlprefix}{URL }
\providecommand{\doi}[1]{https://doi.org/#1}

\bibitem{bochkovskiy2020yolov4}
Bochkovskiy, A., Wang, C.Y., Liao, H.Y.M.: Yolov4: Optimal speed and accuracy
  of object detection (2020)

\bibitem{chen2019hybrid}
Chen, K., Pang, J., Wang, J., Xiong, Y., Li, X., Sun, S., Feng, W., Liu, Z.,
  Shi, J., Ouyang, W., et~al.: Hybrid task cascade for instance segmentation.
  In: Proceedings of the IEEE conference on computer vision and pattern
  recognition. pp. 4974--4983 (2019)

\bibitem{chengwhl20}
Cheng, T., Wang, X., Huang, L., Liu, W.: Boundary-preserving mask r-cnn. In:
  ECCV (2020)

\bibitem{dai2017deformable}
Dai, J., Qi, H., Xiong, Y., Li, Y., Zhang, G., Hu, H., Wei, Y.: Deformable
  convolutional networks. In: Proceedings of the IEEE international conference
  on computer vision. pp. 764--773 (2017)

\bibitem{gupta2019lvis}
Gupta, A., Dollar, P., Girshick, R.: Lvis: A dataset for large vocabulary
  instance segmentation. In: Proceedings of the IEEE Conference on Computer
  Vision and Pattern Recognition. pp. 5356--5364 (2019)

\bibitem{He_2017_ICCV}
He, K., Gkioxari, G., Dollar, P., Girshick, R.: Mask r-cnn. In: Proceedings of
  the IEEE International Conference on Computer Vision (ICCV) (Oct 2017)

\bibitem{he2016deep}
He, K., Zhang, X., Ren, S., Sun, J.: Deep residual learning for image
  recognition. In: Proceedings of the IEEE conference on computer vision and
  pattern recognition. pp. 770--778 (2016)

\bibitem{huang2019mask}
Huang, Z., Huang, L., Gong, Y., Huang, C., Wang, X.: Mask scoring r-cnn. In:
  Proceedings of the IEEE conference on computer vision and pattern
  recognition. pp. 6409--6418 (2019)

\bibitem{Kuznetsova_2020}
Kuznetsova, A., Rom, H., Alldrin, N., Uijlings, J., Krasin, I., Pont-Tuset, J.,
  Kamali, S., Popov, S., Malloci, M., Kolesnikov, A., et~al.: The open images
  dataset v4. International Journal of Computer Vision  \textbf{128}(7),
  1956–1981 (Mar 2020). \doi{10.1007/s11263-020-01316-z},
  \url{http://dx.doi.org/10.1007/s11263-020-01316-z}

\bibitem{li2020overcoming}
Li, Y., Wang, T., Kang, B., Tang, S., Wang, C., Li, J., Feng, J.: Overcoming
  classifier imbalance for long-tail object detection with balanced group
  softmax. In: Proceedings of the IEEE/CVF Conference on Computer Vision and
  Pattern Recognition. pp. 10991--11000 (2020)

\bibitem{lin2017feature}
Lin, T.Y., Doll{\'a}r, P., Girshick, R., He, K., Hariharan, B., Belongie, S.:
  Feature pyramid networks for object detection. In: Proceedings of the IEEE
  conference on computer vision and pattern recognition. pp. 2117--2125 (2017)

\bibitem{lin2014microsoft}
Lin, T.Y., Maire, M., Belongie, S., Hays, J., Perona, P., Ramanan, D.,
  Doll{\'a}r, P., Zitnick, C.L.: Microsoft coco: Common objects in context. In:
  European conference on computer vision. pp. 740--755. Springer (2014)

\bibitem{liu20201st}
Liu, Y., Song, G., Zang, Y., Gao, Y., Xie, E., Yan, J., Loy, C.C., Wang, X.:
  1st place solutions for openimage2019 -- object detection and instance
  segmentation (2020)

\bibitem{milletari2016v}
Milletari, F., Navab, N., Ahmadi, S.A.: V-net: Fully convolutional neural
  networks for volumetric medical image segmentation. In: 2016 fourth
  international conference on 3D vision (3DV). pp. 565--571. IEEE (2016)

\bibitem{tan2020equalization}
Tan, J., Wang, C., Li, B., Li, Q., Ouyang, W., Yin, C., Yan, J.: Equalization
  loss for long-tailed object recognition. In: Proceedings of the IEEE/CVF
  Conference on Computer Vision and Pattern Recognition. pp. 11662--11671
  (2020)

\bibitem{zhang2020resnest}
Zhang, H., Wu, C., Zhang, Z., Zhu, Y., Zhang, Z., Lin, H., Sun, Y., He, T.,
  Mueller, J., Manmatha, R., Li, M., Smola, A.: Resnest: Split-attention
  networks (2020)

\end{thebibliography}
\end{document}